\documentclass{article}

\usepackage{microtype}
\usepackage{graphicx}
\usepackage{booktabs}
\usepackage{hyperref}

\usepackage[accepted]{icml2026}

\usepackage{amsmath}
\usepackage{amssymb}
\usepackage{mathtools}
\usepackage{amsthm}
\usepackage[textsize=tiny]{todonotes}
\usepackage[capitalize,noabbrev]{cleveref}
\usepackage{xcolor}

\theoremstyle{plain}

\theoremstyle{definition}

\theoremstyle{remark}

\icmltitlerunning{Signal-Driven Observation for Long-Horizon Web Agents}

\begin{document}

\twocolumn[
\icmltitle{Signal-Driven Observation for Long-Horizon Web Agents}

\icmlsetsymbol{equal}{*}

\begin{icmlauthorlist}
\icmlauthor{Shubham Gaur}{ucsc}
\icmlauthor{Ian Lane}{ucsc}
\end{icmlauthorlist}

\icmlaffiliation{ucsc}{Department of Computer Science and Engineering,
University of California, Santa Cruz, CA, USA}

\icmlcorrespondingauthor{Shubham Gaur}{sgaur2@ucsc.edu}

\icmlkeywords{web agents, observation compression, failure modes,
long-horizon tasks, context degradation, signal detection,
recursive language models, DOM observation, trace diagnostics,
agentic AI}

\vskip 0.3in
]

\printAffiliationsAndNotice{}

\begin{abstract}
Web agents operating over long horizons ingest raw DOM and accessibility trees — routinely tens of thousands of tokens — at every action step, causing progressive context degradation that erodes reasoning well before tasks complete. We argue that this coupling of observation frequency to action frequency is an architectural mistake. Drawing on the insight from Recursive Language Models that querying a document outperforms reading it wholesale, we propose \textit{Signal-Driven Observation (SDO)}: a dedicated sub-call reads the full DOM but returns only task-relevant elements and their selectors, and is re-invoked only when a lightweight signal detector fires — triggered by URL transitions, newly visible interactive elements, action failures, or exogenous browser events. We outline the open problems SDO introduces and call on the community to treat observation compression as a core architectural decision in web agent design.
\end{abstract}

\section{Introduction}
\label{sec:intro}

Web agents are increasingly deployed on long-horizon tasks---booking
travel, completing multi-page forms, navigating enterprise
workflows---yet they fail at rates that make autonomous operation
impractical. In WebArena \citep{DBLP:journals/corr/abs-2307-13854}, even frontier agents achieve roughly 50\%
success, with the majority failures occurring mid-task rather than
at the outset~\citep{DBLP:journals/corr/abs-2603-19685}. The pattern is consistent
across benchmarks: agents get trapped in action loops, lose track of
their original objective, and produce increasingly incoherent behavior
as interaction history accumulates~\citep{DBLP:journals/corr/abs-2512-04307}.

The community has largely attributed these failures to context length
limitations and responded with longer windows, summarization
pipelines, and memory modules. We argue that this diagnosis is incomplete.
The problem is not that agents run out of context. The problem is that
agents are \emph{architecturally required to ingest everything they
observe}, whether or not it is relevant to deciding the next action.

Consider a concrete example. A web agent at step $t$ receives the raw
accessibility tree of the current page---often 20,000 to 80,000
tokens---a screenshot, and its full interaction history. It then
invokes a complete LLM forward pass to decide a single action that may
be as simple as \texttt{fill(\#email, "john@example.com")}. The
information required to make that decision is a single fact: the email
field is focused and empty. The remaining tens of thousands of tokens
are not merely wasted---they actively degrade the model's ability to
reason about what matters. This is not a context length problem. It is
an observation problem.

We call this failure mode \emph{observation over-ingestion}: the
architectural coupling of observation frequency to action frequency,
forcing the agent to re-read the full page state at every step
regardless of whether that state has meaningfully changed. Observation
over-ingestion is a triggering precondition for three well-documented
downstream failures. First, \emph{context rot}---the progressive
degradation of reasoning quality as irrelevant observation tokens
accumulate in the context window. Second, \emph{loop-trapping}---the
agent repeats the same action sequence because its bloated context
prevents it from recognizing it has visited this state before. Third,
\emph{goal drift}---the agent's original objective is buried under
layers of accumulated DOM noise, and it begins pursuing a different,
emergent sub-goal~\citep{DBLP:journals/corr/abs-2505-02709, DBLP:journals/corr/abs-2603-03258}.

This failure mode is reproducible and independent of the model capability.
It occurs in frontier models with 200K-token windows just as it occurs
in smaller models, because the issue is not capacity, but
architecture~\citep{DBLP:journals/corr/abs-2510-24699}. It is also invisible to standard
evaluation: terminal success metrics record that the agent failed but
not that observation over-ingestion was the mechanism. Trace-level
diagnostics that could surface this failure---logging when the
observation changed versus when the agent re-read it
unchanged---do not exist in current evaluation frameworks.

Recursive Language Models~(RLMs)~\citep{DBLP:journals/corr/abs-2512-24601} offer an
architectural insight that has not yet been applied to this problem.
RLMs demonstrated that for large static documents, models perform
dramatically better when they query a document programmatically rather
than ingesting it wholesale. The model treats the document as an
external variable, calling sub-processes to extract exactly what it
needs. The live DOM of a web page is precisely the environment where
this insight is most needed---and most absent.

We sketch one concrete instantiation of this principle---\emph{Signal-Driven
Observation}~(SDO)---not as a complete system but to demonstrate that
observation over-ingestion is architecturally avoidable and to surface
the open design questions a solution must address. In SDO, a dedicated
sub-call reads the full DOM but returns only a compact,
task-conditioned summary. This sub-call is not invoked at every
action step. A lightweight signal detector monitors four browser-native
events---URL transitions, newly visible interactive elements, action
failures, and exogenous page events---and triggers re-observation only
when the page state has meaningfully changed. Between signals, the
root model executes its planned action sequence with no additional LLM
calls and no growing context.

This paper makes three contributions. First, we define observation
over-ingestion as an operational failure mode in web agents---a
reproducible triggering precondition for context rot, loop-trapping,
and goal drift that is distinct from context length exhaustion. Second,
we sketch Signal-Driven Observation as a concrete demonstration that
observation frequency can be decoupled from action frequency. Third,
we identify the open problems this framing surfaces and call on the
community to treat observation compression as a first-class target for
failure mitigation in agentic systems.

\section{Related Work}
\label{sec:related}

\subsection{Web Agent Benchmarks and Long-Horizon Evaluation}

The evaluation landscape for web agents has expanded rapidly.
WebArena~\citep{DBLP:journals/corr/abs-2307-13854} introduced self-hosted, reproducible
web environments with programmatic evaluation;
WorkArena~\citep{DBLP:journals/corr/abs-2403-07718}, WorkArena++ \citep{DBLP:journals/corr/abs-2407-05291} extended this to enterprise
ServiceNow workflows where single HTML pages can reach
40K--500K tokens. BrowserGym~\citep{DBLP:journals/corr/abs-2412-05467} provides the
unified observation framework underlying several of these
benchmarks, exposing the full accessibility tree, raw HTML, and
viewport screenshots at every step---a single observation routinely
exceeding 20,000 tokens.
VisualWebArena~\citep{DBLP:journals/corr/abs-2401-13649} and
OSWorld~\citep{DBLP:journals/corr/abs-2404-07972} broadened evaluation to multimodal and
desktop-level tasks respectively.
More recently, Online-Mind2Web~\citep{DBLP:journals/corr/abs-2504-01382}
demonstrated that frontier agents are up to 59\% less competent on
live websites than static benchmarks suggest, attributing much of
the gap to dynamic content and evaluation artifacts.
REAL~\citep{DBLP:journals/corr/abs-2504-11543} provides deterministic website replicas for
reproducible evaluation, while Odysseys~\citep{jang2026odysseysbenchmarkingwebagents}
introduces 200 long-horizon, multi-site tasks derived from real
browsing histories and explicitly identifies the inadequacy of
trajectory-level LLM-as-judge evaluation for long tasks.

The pattern across these benchmarks is consistent: as task horizon
grows, agent success degrades sharply.
HORIZON~\citep{DBLP:journals/corr/abs-2604-11978} formalizes this with a cross-domain
diagnostic benchmark and trajectory-grounded failure attribution,
showing that long-horizon failures are systematically
under-characterized by terminal metrics.
\citep{DBLP:journals/corr/abs-2512-04307} find that agents primarily fail by getting
stuck in loops and losing track of objectives---not from exhausting
context tokens. A subgoal-driven analysis~\citep{DBLP:journals/corr/abs-2603-19685}
reports mid-task stuck behavior in nearly 50\% of WebArena-Lite
trajectories for Gemini-2.5-Pro. These findings motivate our claim
that the failure mechanism lies in \emph{how} agents observe, not
\emph{how much} they can store.

\subsection{Context and Observation Management}

A growing body of work addresses the observation burden directly.
AgentFold~\citep{DBLP:journals/corr/abs-2510-24699} reports that approximately 20\%
of long-horizon tasks are forcibly terminated at 100 turns despite
contexts using only around 7K tokens, suggesting failures arise
from in-context confusion rather than raw length.
FocusAgent~\citep{DBLP:journals/corr/abs-2510-03204} uses a lightweight LLM retriever
to extract task-relevant lines from accessibility-tree observations,
reducing observation size by over 50\% on WorkArena and WebArena
while also lowering prompt-injection success rates.
LineRetriever~\citep{DBLP:journals/corr/abs-2507-00210} proposes planning-aware
observation reduction via embedding-based retrieval, explicitly
motivated by the insight that agents ``need future-action-relevant
context, not just semantically similar text.''
ACON~\citep{DBLP:journals/corr/abs-2510-00615} provides a unified framework that compresses
both observations and interaction histories through failure-driven
natural-language guidelines, achieving 26--54\% peak-token reduction
on AppWorld \citep{DBLP:journals/corr/abs-2407-18901} and OfficeBench \citep{DBLP:journals/corr/abs-2407-19056}.
Hierarchical Memory Tree~\citep{DBLP:journals/corr/abs-2603-07024} abstracts raw HTML
trajectories into compact semantic descriptions, reporting 72.7\%
context-length reduction on WebArena.
SLIM~\citep{yen2025lostmazeovercomingcontext} introduces summarization for accumulated
search content, dual-memory frameworks such as
M\textsuperscript{2}~\citep{DBLP:journals/corr/abs-2603-00503} separate working from
long-term memory, and ContextBudget~\citep{DBLP:journals/corr/abs-2604-01664}
introduces dynamic budget-conditioned compression that explicitly
identifies failure modes of budget-free approaches.

A counterpoint is offered by \citep{DBLP:journals/corr/abs-2604-01535}, who
argue that the optimal observation representation depends on model
capability---compact accessibility trees for weaker models, raw
HTML with thinking budget for stronger ones. This highlights that
compression is not uniformly beneficial, a tension our work
acknowledges.

All of these approaches share a common assumption: the agent
\emph{should} observe the full page state and the engineering
problem is compressing it afterward. None questions whether full
observation should occur at every step in the first place.
Observation over-ingestion is treated as a data-volume problem
rather than a frequency problem.

\subsection{Failure Diagnosis, Safety, and Trace-Level Attribution}

The failure-diagnosis literature has matured rapidly. On the
robustness side, WAREX~\citep{DBLP:journals/corr/abs-2510-03285} injects network errors,
server failures, and malicious popups into WebArena and REAL,
demonstrating significant task-success drops under realistic
perturbations. StressWeb~\citep{bai2026stresswebdiagnosticbenchmarkweb} introduces controlled
perturbations across the perception, semantic, and execution stages
of the interaction pipeline.
DoomArena~\citep{DBLP:journals/corr/abs-2504-14064} attacks BrowserGym agents with
popup injections hidden in accessibility attributes, achieving
23--78\% failure rates on OSWorld.
\citep{DBLP:journals/corr/abs-2411-02391} show that adversarial popups alone reach
86\% click-through rates and reduce task success by 47\%.

The safety implications of accumulated context are also emerging:
MT-AgentRisk~\citep{DBLP:journals/corr/abs-2602-13379} shows that multi-turn observation
histories create attack surfaces beyond the capability failures we
focus on here.

On the attribution side, AgentRx~\citep{DBLP:journals/corr/abs-2503-18102} provides a
nine-category failure taxonomy over 115 annotated failed
trajectories. AgenTracer~\citep{DBLP:journals/corr/abs-2509-03312} uses counterfactual
replay and programmed fault injection.
ST-WebAgentBench~\citep{DBLP:journals/corr/abs-2410-06703} introduces
Completion-under-Policy, showing that safety-adjusted success can
be less than two-thirds of nominal completion rates.
\citep{DBLP:conf/emnlp/FangZG25} study preemptive detection and
correction of misaligned actions in web shopping agents---exactly
the kind of grounding error that bloated observations make harder
to catch.

These frameworks diagnose failures at the \emph{action} level:
which action was wrong, which policy was violated, which step was
the root cause. None diagnoses failures at the
\emph{observation} level---whether the agent re-ingested an
unchanged page, whether a fresh observation would have prevented
the error, or whether the volume of observation tokens was itself
the degradation mechanism. If observation over-ingestion is the
triggering precondition, then action-level attribution will
consistently mislocate the cause.

\subsection{Recursive and Hierarchical Agent Architectures}

Recursive Language Models~(RLMs)~\citep{DBLP:journals/corr/abs-2512-24601} treat the input
prompt as an external variable in a REPL environment, recursively
calling sub-instances to peek at relevant slices without bloating
the root context. On long-context benchmarks, RLMs maintain strong
performance at inputs exceeding ten million tokens where standard
models degrade sharply. However, RLMs were designed for static
documents---codebases, legal corpora, long-form text. The live DOM
of a web page mutates with every agent action and can be
interrupted by exogenous events independent of the agent's
behavior. Applying recursive decomposition to a dynamic, stateful
environment requires addressing when to re-observe and what to
extract, neither of which arises in the static setting.

Prior web agent architectures have explored related structural
ideas without addressing observation frequency.
LASER~\citep{DBLP:journals/corr/abs-2309-08172} introduces URL-based state tracking with
backtracking on e-commerce tasks.
BEAP-Agent~\citep{DBLP:journals/corr/abs-2601-21352} uses depth-first search with
multi-level backtracking.
RCI~\citep{DBLP:journals/corr/abs-2303-17491} adds self-critique loops that evaluate action
quality.
RAGEN~\citep{DBLP:journals/corr/abs-2504-20073} introduces StarPO for training agentic
reasoning and documents the ``Echo Trap'' failure mode---agents
echoing earlier thoughts---which is an observation-context
pathology directly caused by accumulated context noise.
Subgoal-driven frameworks~\citep{DBLP:journals/corr/abs-2603-19685} decompose
long-horizon tasks into explicit sub-plans.
All of these operate \emph{over} the observation---planning what
to do with what the agent sees---but none intervenes on the
observation itself. At every action step, the full DOM is still
read, the full context still grows, and the same over-ingestion
mechanism still applies.

The gap is precise: no existing architecture decouples observation
frequency from action frequency in web agents. We propose
Signal-Driven Observation as one concrete way to close it.

\section{Signal-Driven Observation}
\label{sec:sdo}

\begin{figure*}[t]
  \centering
  \includegraphics[width=\textwidth]{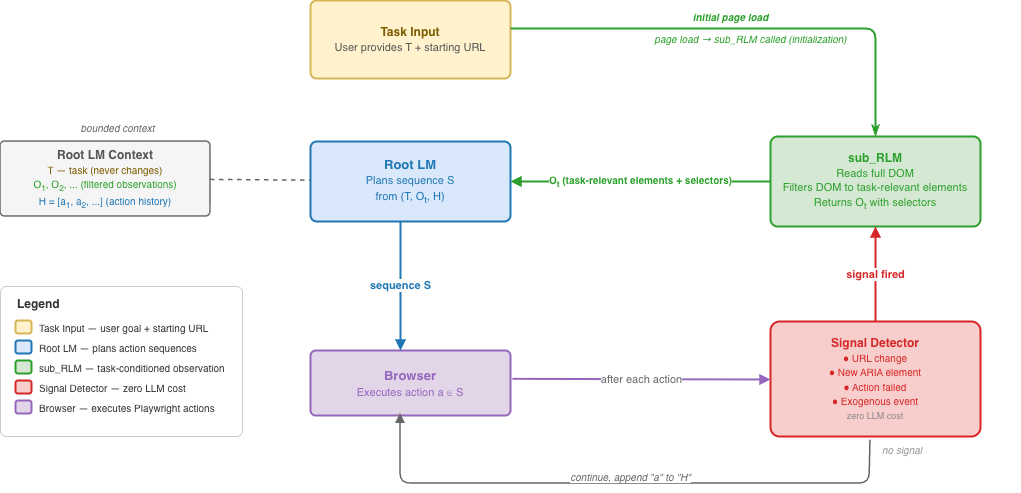}
  \caption{SDO architecture. The Signal Detector runs 
  after every action at zero LLM cost. sub\_RLM is 
  invoked only when a signal fires, returning a compact 
  observation $O_{t+1}$. The Root LM replans from 
  bounded context.}
  \label{fig1}
\end{figure*}

We sketch \emph{Signal-Driven Observation}~(SDO)
(Figure~\ref{fig1})---not as a complete system but as a concrete
instantiation of a principle: observation frequency should be
decoupled from action frequency. SDO makes this precise by defining
a set of lightweight, zero-LLM-cost signals that determine when a
new observation is warranted, and delegating all DOM reading to a
focused sub-call that the root LM never performs itself. We frame
each architectural component in terms of the failure mode it is
designed to prevent.

\subsection{Architecture}

SDO involves four components operating at runtime over a standard
browser controlled via Playwright.

\paragraph{Root LM.}
The root LM maintains three variables throughout the task: the
original task specification~$T$, which never changes; a sequence of
compact observations $O_1, O_2, \ldots$ received from sub\_RLM;
and an action history $H = [a_1, a_2, \ldots]$ that grows by one
entry after every executed action. The root LM plans a sequence of
actions $S$ from $(T, O_t, H)$ and replans only when a new
observation arrives.

The failure-mode logic is direct. Because the root LM never reads
raw DOM, its context contains only task-relevant information:
compact observations and single-line action entries. The
signal-to-noise ratio in the root LM's context remains high
throughout the task, removing the primary triggering precondition
for context rot. And because $T$ is never buried under thousands
of irrelevant DOM tokens, the root LM's access to the original
task objective is preserved, reducing the conditions under which
goal drift emerges.

\paragraph{sub\_RLM.}
sub\_RLM is an LM call---either the same model as the root LM
under a different prompt, or a smaller dedicated model---whose sole
responsibility is DOM observation. Given the current page and the
task $T$, it reads the full accessibility tree and returns a
compact, task-conditioned observation $O_t$: a structured list of
only the task-relevant interactive elements, each described by its
purpose, type, current value, and Playwright selector. sub\_RLM
does not plan or execute actions. It answers one question:
\emph{what on this page matters for the current task, and where
is it?} We describe the text-only case for concreteness, but
sub\_RLM can equally operate over a screenshot, or both modalities
together---the architectural contribution is the decoupling of
observation from action, not the specific modality of the sub-call.

The task-conditioning is critical for failure prevention. A raw
accessibility tree contains every element on the page---navigation
menus, footer links, sidebar widgets, hidden elements, advertising
containers. Each of these is noise that competes with the
task-relevant signal. By filtering to only elements that matter for
$T$, sub\_RLM removes this noise at the source rather than asking
the root LM to filter it under growing context pressure.

\paragraph{Signal Detector.}
The signal detector runs after every Playwright action at zero LLM
cost. It monitors four browser-native conditions:

\begin{enumerate}
    \item \textbf{URL transition}---the page navigated to a new
    address, indicating a state change that invalidates the
    current observation.
    \item \textbf{New ARIA element}---an element with a role such
    as \texttt{listbox}, \texttt{select}, \texttt{menu}, \texttt{combobox}, 
    \texttt{dialog}, or with \texttt{aria-expanded=true} became 
    visible that was not present in $O_t$. This covers dropdowns opening, modals appearing, and
    dynamic widgets loading---all cases where new interactive
    content that the root LM needs to act on has appeared in the
    DOM.
    \item \textbf{Action failure}---the Playwright action raised an
    exception, timed out, or the targeted element became
    unavailable, indicating the page is not in the state the root
    LM assumed.
    \item \textbf{Exogenous event}---a DOM mutation occurred that
    was not caused by the agent's action: a cookie consent banner
    injected by a JavaScript timer, an AJAX response rewriting page
    content, a session timeout redirect, or a third-party script
    modifying layout. Signals~1--3 arise in both simulated
    environments and live browsers. Signal~4 arises primarily on
    real websites---a gap we discuss in
    Section~\ref{sec:openproblems}.
\end{enumerate}

Beyond its role in triggering re-observation, the signal detector
produces a structured trace of environment state changes across the
trajectory. This trace---which signals fired at which steps, and of
which type---is itself a diagnostic artifact that current evaluation
frameworks do not capture. It enables a form of observation-level
attribution: when the agent fails at step $k$, the trace can reveal
whether a signal was missed at an earlier step, whether the
observation was stale, or whether a re-observation was triggered
but produced an insufficient $O_t$.

\paragraph{Browser.}
The browser executes Playwright actions from the sequence $S$
produced by the root LM. After each action, control passes to the
signal detector before the next action in $S$ begins. The root LM
never interacts with the browser directly.

\subsection{The SDO Loop}

On task start, the initial page load triggers a sub\_RLM call,
producing the first observation $O_1$ before the root LM has taken
any action. The root LM receives $(T, O_1, H{=}[])$ and plans an
initial action sequence $S$. For each action $a \in S$: the
browser executes $a$; the action is appended to $H$; the signal
detector checks the four conditions. If no signal fires, the next
action in $S$ executes immediately---no LLM call, no DOM read. If
any signal fires, sub\_RLM is called on the current DOM, producing
$O_{t+1}$; the root LM receives the updated $(T, O_{t+1}, H)$ and
replans from the new observation.

\subsection{Case Study: E-Commerce Purchase}

We trace a concrete long-horizon task---\emph{``Find a pair of
wireless noise-cancelling headphones under \$200 on an e-commerce
site and complete checkout''}---to illustrate both how observation
over-ingestion causes cascading failures under the standard
approach and how SDO prevents them.

\paragraph{Standard agent (without SDO).}

The agent lands on the homepage. The raw DOM contains approximately
1,200 elements: navigation bars, promotional carousels, trending
product grids, footer links, search bars, and login prompts. The
agent ingests all of it and decides to search for ``wireless
noise-cancelling headphones.''

\emph{Step 1--3: Search and filter.} The agent types the query and
submits. The results page contains 48 product listings, sponsored
results, filter panels, pagination controls, and a sorting
dropdown. Each product listing includes title, price, rating,
review count, and availability. The agent ingests all of it,
identifies a promising pair at \$179, and clicks through.

\emph{Step 4--6: Product detail page.} The product page includes
specifications, customer reviews, related products, ``frequently
bought together'' suggestions, seller information, and shipping
options. By this point the root LM's context contains the homepage
DOM, the results page DOM, and the product page DOM---three full
raw snapshots, each tens of thousands of tokens, accumulating into
a context dominated by irrelevant page content. A
cookie consent banner fires (exogenous event). The banner overlays
the ``Add to Cart'' button. The agent, whose context is already
degraded, clicks the banner's ``Accept'' button but interprets the
resulting page state incorrectly---it believes it has added the
item to cart.

\emph{Step 7--10: Cart and checkout.} The agent navigates to the
cart. The cart is empty---the item was never added. But the agent's
context now spans four full DOM snapshots, each tens of thousands of
tokens. The original task specification---\emph{wireless
noise-cancelling headphones under \$200}---is buried deep in the
context. The agent sees ``Your cart is empty'' but, rather than
recognizing the failure and backtracking, it searches for a new
product (goal drift). It finds a \$350 pair of wired headphones---
violating both the wireless and the price constraints. By step 12,
it has begun the checkout process for the wrong product. Terminal
evaluation records: \emph{task failed.} It does not record where or
why.

\paragraph{SDO agent.}

The agent lands on the same homepage. sub\_RLM reads the 1,200
elements and returns $O_1$: a single task-relevant element---the
search bar, with its selector and purpose. The root LM's context
contains the task and one compact observation. It plans: type query,
click search.

\emph{Step 1--3: Search and filter.} After submitting the search,
the URL changes---signal fires. sub\_RLM reads the results page.
48 listings exist, but only 6 are wireless noise-cancelling
headphones under \$200. $O_2$ contains those 6, each with name,
price, and selector---a compact representation orders of magnitude
smaller than the raw results page. It selects the \$179 pair and
clicks.

\emph{Step 4--6: Product detail page.} URL change---signal fires.
sub\_RLM returns $O_3$: product name, price (\$179), ``Add to
Cart'' button with selector, and one unexpected element---a cookie
consent banner flagged as \texttt{task\_relevant: false,
blocking: true}. Root LM sees this is an interruption, not a state
change. It plans: dismiss banner, then click Add to Cart. Two
actions, no LLM call between them. After ``Add to Cart,'' a small
AJAX confirmation appears (new ARIA element---signal fires).
sub\_RLM confirms: ``Item added. Cart count: 1.'' The root LM's
context has grown only by the addition of compact observations and
action entries---not by accumulated raw DOM.

\emph{Step 7--10: Cart and checkout.} The agent navigates to cart.
URL change---signal fires. $O_5$: cart contains ``Sony WH-1000XM5,
\$179, quantity 1.'' Root LM compares against $T$: wireless
noise-cancelling, under \$200---match. Proceeds to checkout.
Each checkout page triggers one sub\_RLM call, returning only the
form fields needed. The agent completes checkout in 4 actions
across 2 pages.

\emph{Outcome:} 12 actions, 6 sub\_RLM calls. The root LM's context
grew only at signal boundaries---each time by a compact observation
of a few task-relevant elements---rather than by tens of thousands
of raw DOM tokens at every step. The task specification $T$ remained
immediately accessible throughout the trajectory, never buried under
irrelevant page content.
No context rot. No goal drift. The cookie banner was handled as an
interruption, not a confusion point.

\paragraph{Failure-mode comparison.}

The standard agent failed not because any individual action was
wrong, but because accumulated observation noise degraded its
ability to maintain task coherence. The cookie banner was the
proximate trigger, but the precondition was multiple full DOM
snapshots of irrelevant content already in context. SDO prevents the precondition:
the root LM never sees irrelevant DOM, so when the cookie banner
appears, it arrives in a clean context where the root LM can
correctly classify it as an interruption and plan around it.

This illustrates the core claim: observation over-ingestion is not
merely inefficient---it is a failure mechanism. Removing it does
not just save tokens. It preserves the conditions under which the
root LM can reason correctly about the task, even when the
environment introduces unexpected complications.

\section{Open Problems}
\label{sec:openproblems}

SDO is a sketch, not a solution. It demonstrates that observation
over-ingestion is architecturally avoidable, but it surfaces a set
of design questions that must be resolved before the principle can
be instantiated reliably. We organize these as open problems for
the community, each with an explicit description of the tradeoff
involved.

\paragraph{Signal completeness versus cost.}
The four signals we define---URL transition, new ARIA element,
action failure, and exogenous event---are lightweight and cover
the most common state changes in browser environments. But they
are not exhaustive. A page can change meaningfully without
triggering any of these: a price updating silently via JavaScript,
a product going out of stock with no visible DOM change, a form
validation message appearing inside an existing element rather than
as a new one. Missing such changes means the root LM plans against
a stale observation. Adding more signals increases coverage but
also increases the frequency of sub\_RLM calls, eroding the
efficiency gain that motivates SDO in the first place. The
tradeoff is between observation freshness and call frequency, and
the right balance likely depends on the task domain and the cost
of acting on stale information.

\paragraph{Task-conditioned filtering fidelity.}
sub\_RLM must decide which elements are task-relevant without
knowing what the root LM will need. If it filters too aggressively,
it omits elements the root LM would have used---a form of
information loss that is unrecoverable without a full
re-observation. If it filters too conservatively, observations
grow large and the compression benefit diminishes. In structured
environments like WorkArena, where task parameters map directly to
form fields, the filtering problem is tractable. On open-ended
real-web tasks---\emph{``find a good anniversary gift under
\$100''}---task relevance is ambiguous and sub\_RLM must make
judgment calls that may not match the root LM's reasoning. This
is a compression-completeness tradeoff: how much information can
be discarded at the observation layer without degrading the
planning layer?

\paragraph{Semantic errors within a state.}
SDO's signal detection catches structural and technical failures
but not semantic ones. If the root LM plans the wrong action---
selecting ``Software'' when the task requires ``Hardware''---and
the action executes successfully with no structural change to the
page, no signal fires. The agent continues on the wrong path
without any observation-level indication that something went
wrong. This error is only caught downstream: either when a later
sub\_RLM call reveals an inconsistency, or at terminal evaluation
when the task is already failed. A natural extension is a
lightweight value check after form-filling actions---comparing the
executed value against the planned value via a direct DOM
read---but this check is reliable only in environments with stable
selectors and standard HTML form elements. On real websites using
React-controlled inputs or custom components, the DOM value and
the application state can diverge. Detecting mid-sequence semantic
drift without structural signals remains an open problem.

\paragraph{Simulation versus real-web gap.}
Signals 1--3 (URL transition, new ARIA element, action failure)
arise naturally in both simulated benchmarks and live browsers.
Signal~4 (exogenous events) exposes a fundamental gap: simulated
environments such as WebArena and WorkArena are deterministic by
design---no cookie banners, no AJAX surprises, no session
timeouts. SDO's signal detector can capture exogenous events on
live browsers using standard Playwright event listeners and DOM
mutation observers, requiring no model training. However, this
gap has broader implications for the community: agents that are
trained via reinforcement learning in deterministic simulation
never encounter exogenous events during training and therefore
have no learned recovery behavior when deployed on real
websites. SDO
sidesteps this for inference---the signal detector is
rule-based, not learned---but whether the root LM can
reliably plan around exogenous interruptions it has never
seen in its pretraining or prompting context remains an
open question.

\paragraph{Observation-level evaluation.}
Current web agent benchmarks evaluate at the action level (was the
action correct?) or the task level (did the agent succeed?). SDO
motivates a third level: \emph{observation-level evaluation}. Was
the observation fresh when the agent needed it? Did the agent act
on stale information? Would a re-observation at step $k$ have
prevented the failure at step $k{+}3$? The signal detector's trace
provides the raw data for this kind of analysis, but the metrics
and benchmarks that would operationalize it do not yet exist.
Building them is a prerequisite for measuring whether observation
compression actually prevents the failures it is designed to
address, rather than merely reducing token count.

\paragraph{Observation history management.}
As the trajectory lengthens, the list of past observations
$O_1, O_2, \ldots, O_k$ grows in the root LM's context. While
each individual observation is compact, the cumulative history can
still become substantial over very long horizons---tasks spanning
dozens of pages and hundreds of actions. At some point, older
observations may need to be summarized or discarded, reintroducing
a version of the context management problem SDO was designed to
avoid. How to manage observation history without losing information
needed for backtracking or loop detection is an open design
question.

\section{Conclusion}
\label{sec:conclusion}

We have argued that observation over-ingestion---the architectural
coupling of observation frequency to action frequency---is a
first-class failure mode in web agents, distinct from context
length exhaustion and serving as a reproducible triggering
precondition for context rot, loop-trapping, and goal drift. We
sketched Signal-Driven Observation as one concrete demonstration
that this coupling is not inevitable: a lightweight signal detector
can govern when re-observation occurs, and a focused sub-call can
compress what the agent sees without the root model ever reading
raw DOM. The approach is a direction, not a finished system, and we
have been explicit about what it cannot yet handle---semantic
errors without structural signals, filtering fidelity on
open-ended tasks, and the absence of exogenous events from
simulated training environments. We also note that this paper 
does not include experiments.
Whether SDO prevents the failures we describe in practice is
a question that can only be answered by building and
evaluating it---a natural next step that we hope this
framing motivates.

The broader point is simple. The community has invested heavily in
making agents reason better over what they observe. Comparatively
little attention has gone to whether agents should be observing
what they observe in the first place. We believe observation
compression deserves to be treated as a first-class design
primitive---not a post-hoc optimization, but a structural decision
made before the first action is taken. We hope this paper
motivates others to formalize the problem and build the evaluation
infrastructure needed to measure whether solving it actually
prevents the failures it is designed to address.

\bibliographystyle{icml2026}
\bibliography{references}




\end{document}